\documentclass[conference]{IEEEtran}
\IEEEoverridecommandlockouts
\usepackage{cite}
\usepackage{amsmath,amssymb,amsfonts}
\usepackage{algorithmic}
\usepackage{graphicx}
\usepackage{textcomp}
\usepackage{xcolor}

\usepackage{amsmath,amssymb,amsfonts}
\usepackage{multirow}
\usepackage{graphicx}
\usepackage{rotating}
\usepackage{multicol}
\usepackage{amsmath}
\usepackage{float}
\usepackage{array}
\usepackage{caption}
\usepackage{subcaption}
\usepackage{siunitx}
\usepackage{float}
\usepackage{xcolor}
\usepackage{multirow}
\usepackage{graphicx}
\usepackage{color, soul, colortbl}
\usepackage{nicematrix}
\usepackage{colortbl} 
\usepackage{hyperref}
\usepackage{algorithm}
\usepackage{algorithmic}
\usepackage{balance}

\DeclareUnicodeCharacter{2061}{}

\def\BibTeX{{\rm B\kern-.05em{\sc i\kern-.025em b}\kern-.08em
    T\kern-.1667em\lower.7ex\hbox{E}\kern-.125emX}}
\begin{document}

\title{ Maritime Mission Planning for Unmanned Surface Vessel using Large Language Model}

\author{Muhayy Ud Din,  Waseem Akram, Ahsan B Bakht, Yihao Dong, and Irfan Hussain$^{*}$                                                     
 \thanks{$^{*}$Khalifa University Center for Advanced Robotic Systems (KUCARS), United Arab Emirates.} 
 \thanks{$^{*}$This work is supported by the Khalifa University of Science and Technology under Award No. 8434000534, CIRA-2021-085,
RC1-2018-KUCARS, KU-Stanford :8474000605.}%
\thanks{$^{*}$ Corresponding Author, Email: irfan.hussain@ku.ac.ae}
}
\maketitle

\begin{abstract}
Unmanned Surface Vessels (USVs) are essential for various maritime operations. USV mission planning approach offers autonomous solutions for monitoring, surveillance, and logistics. Existing approaches, which are based on static methods, struggle to adapt to dynamic environments, leading to suboptimal performance, higher costs, and increased risk of failure.
This paper introduces a novel mission planning framework that uses Large Language Models (LLMs), such as GPT-4, to address these challenges. LLMs are proficient at understanding natural language commands, executing symbolic reasoning, and flexibly adjusting to changing situations. Our approach integrates LLMs into maritime mission planning to bridge the gap between high-level human instructions and executable plans, allowing real-time adaptation to environmental changes and unforeseen obstacles. In addition, feedback from low-level controllers is utilized to refine symbolic mission plans, ensuring robustness and adaptability.
This framework improves the robustness and effectiveness of USV operations by integrating the power of symbolic planning with the reasoning abilities of LLMs. In addition, it simplifies the mission specification, allowing operators to focus on high-level objectives without requiring complex programming. The simulation results validate the proposed approach, demonstrating its ability to optimize mission execution while seamlessly adapting to dynamic maritime conditions.
\end{abstract}


\IEEEpeerreviewmaketitle

\section{Introduction}

Unmanned Surface Vessels (USVs) have become essential in modern maritime operations ~\cite{din2023marine}, offering a cost-effective and adaptable solution to various challenges in the marine environment~\cite{bae2023survey}. From environmental monitoring and surveillance to logistics and security, USVs play a crucial role in tasks where safety concerns, logistical constraints, or operational efficiency limit human involvement. The ability of USVs to operate autonomously in complex and dynamic environments, supported by enhancement techniques~\cite{bakht2024mula}, has opened new possibilities in both civilian and defense applications, highlighting the need for advanced mission planning techniques that can fully utilize their potential~\cite{ahmed2023vision}.

USV mission planning in maritime environments involves designing optimal paths and actions to achieve specific objectives~\cite{chiang2018path}~\cite{heo2017case}, such as surveillance\cite{din2024benchmarking}, data collection\cite{ZHAO2023115781}, or inspection~\cite{akram2022visual} while navigating in complex and dynamic conditions.
It is a fundamental aspect of USV operations and contains detailed strategies to achieve defined goals. Existing USV mission planning approaches mainly rely on rule-based systems~\cite{jin2022cooperative}, heuristic algorithms~\cite{zhao2023path}, or predefined routines~\cite{zhang2024predefined}. While these methods demonstrate potential in controlled environments, they frequently lack the adaptability needed for the dynamic and uncertain nature of maritime operations. For example, rule-based systems often fail when faced with uncertainty of maritime conditions, such as encountering unknown obstacles that lead to mission delays or failures \cite{chiang2018path}. Similarly, heuristic algorithms, despite their computational efficiency, may fail to produce optimal plans in complex, multi-objective scenarios, such as navigation in congested waterways or cooperative missions involving multiple USVs \cite{ZHU2025120165}. These limitations not only reduce mission efficiency, but also increase operational costs and the risk of mission failure, highlighting the need for more robust and adaptive mission planning frameworks \cite{thompson2019review}.

\begin{figure*}[t]
\includegraphics[width=\linewidth]{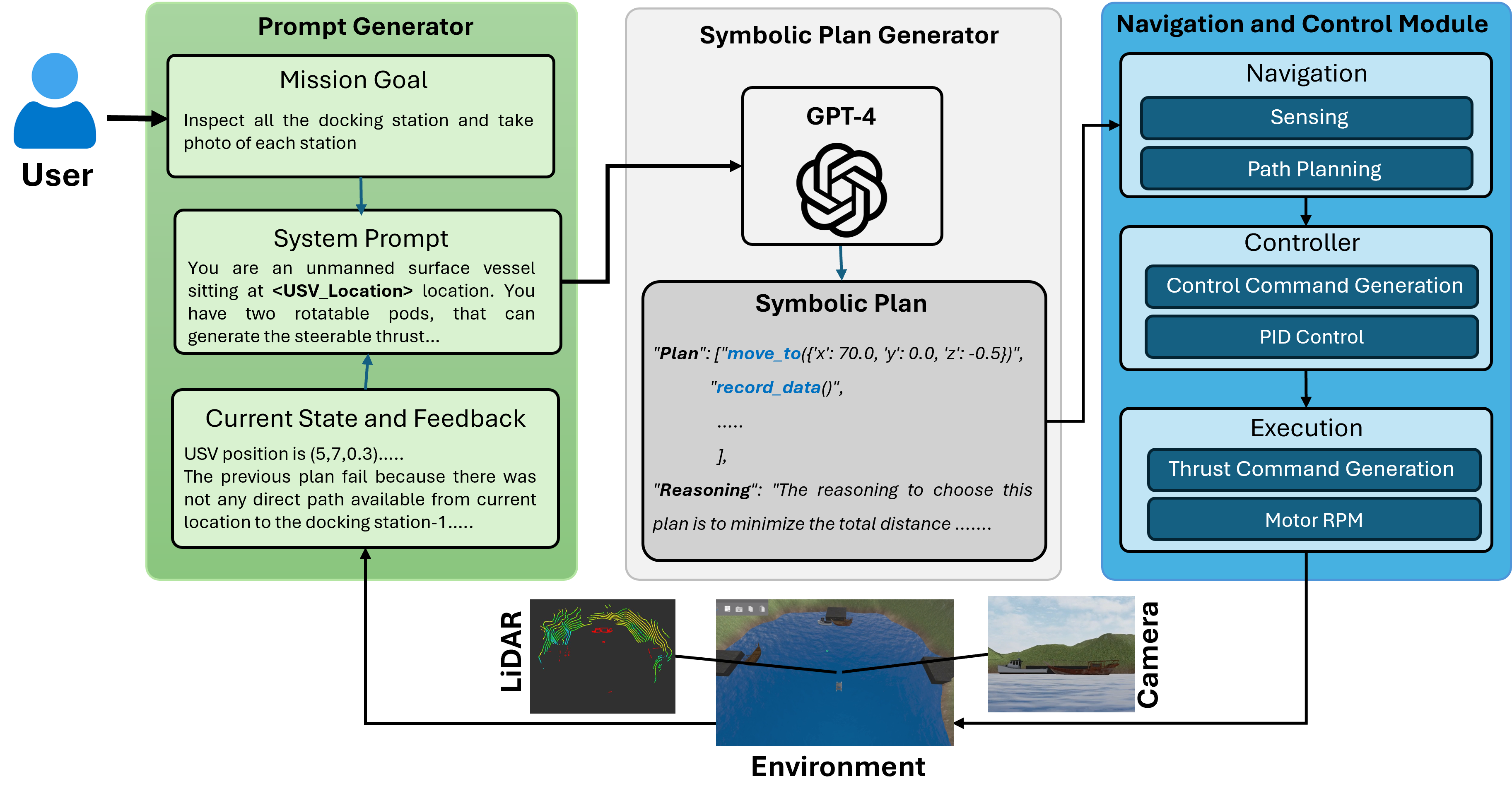}
\caption{The LLM-guided mission planning framework consists of three main modules: the Prompt Generator, which formulates the mission goal and system prompt; the Symbolic Plan Generator, powered by GPT-4, to create an adaptive symbolic plan; and the Navigation and Control Module, which handles sensing, path planning, and precise execution using PID control. Real-time visual outputs provide feedback on perception and task performance.}
\label{fig:framework11}
\end{figure*}

This paper addresses these challenges by introducing a novel approach to mission planning for USVs using large language models (LLM) such as GPT-4 ~\cite{achiam2023gpt}.
LLMs represent cutting-edge AI technology that is adept at both interpreting and producing language that mimics human-like interactions~\cite{brown2020language}. Their capability in handling complex instructions, performing symbolic reasoning, and improving decision making make them a highly suitable choice to advance planning in USV missions.
Using advanced natural language understanding and contextual reasoning, LLMs can dynamically interpret mission requirements, assess environmental changes, and propose adaptive plans in real-time ~\cite{wang2024survey}.


Our approach utilizes LLMs to generate structured sequences of actions designed to achieve mission goals efficiently ~\cite{zhu2024language}. For example, consider a scenario where a USV visits four docking terminals to capture images. An LLM can dynamically compute a symbolic plan that prioritizes the nearest terminal first, optimizes the travel distance, and minimizes the mission time. However, if there is an obstacle in the way that blocks direct access to the nearest terminal, the low-level control will feedback the reason of failure to the LLMs, which will regenerate the symbolic plan by overcoming the reason of failure. 

The core contributions of this work are as follows:

\begin{itemize}
\item \textit{Integration of LLMs into Maritime Mission Planning:} We demonstrate the application of Large Language Models in USV mission planning, bridging the gap between natural language instructions and executable plans.

\item \textit{Optimized and Adaptive Symbolic Planning:} Our approach ensures that mission objectives, such as visiting specified locations or collecting data, are achieved efficiently and with a goal-oriented approach. Account for both mission constraints and dynamic environmental factors.

 \item \textit{Real-Time Plan Adaptation with Feedback Integration:} By utilizing the reasoning and analytical capabilities of LLMs, our framework enables USVs to dynamically adapt mission plans in response to real-time changes, such as route blockages or altered priorities. In addition, low-level controller feedback is incorporated to refine symbolic mission plans.

\end{itemize}

\section{Proposed Approach}

The proposed framework, illustrated in Fig.~\ref{fig:framework11}, is designed to enable an Unmanned Surface Vessel to autonomously plan and execute missions in a maritime environment. To validate the effectiveness of the framework, we applied it to a mission focused on inspecting docking terminals and capture photographs of each terminal. This mission demonstrates the framework's capability to integrate high-level symbolic reasoning with low-level control and execution in a complex operational setting.
It is composed of three primary components:

\begin{enumerate}
    \item \textit{Prompt Generator}: Responsible for dynamically generating mission-specific prompts that incorporate task instructions, environmental data, and feedback from the environment.
    \item \textit{LLM-Based Symbolic Plan Generator}: It uses the capabilities of a Large Language Model to produce high-level symbolic plans that align with the mission objectives and environmental constraints.
    \item \textit{Low-Level Navigation and Control Module}: Ensures precise execution of the generated plans by managing the USV's real-time navigation and control systems.
\end{enumerate}
The detailed description of these components is given below.

\begin{figure}[t]
\includegraphics[width=\columnwidth]{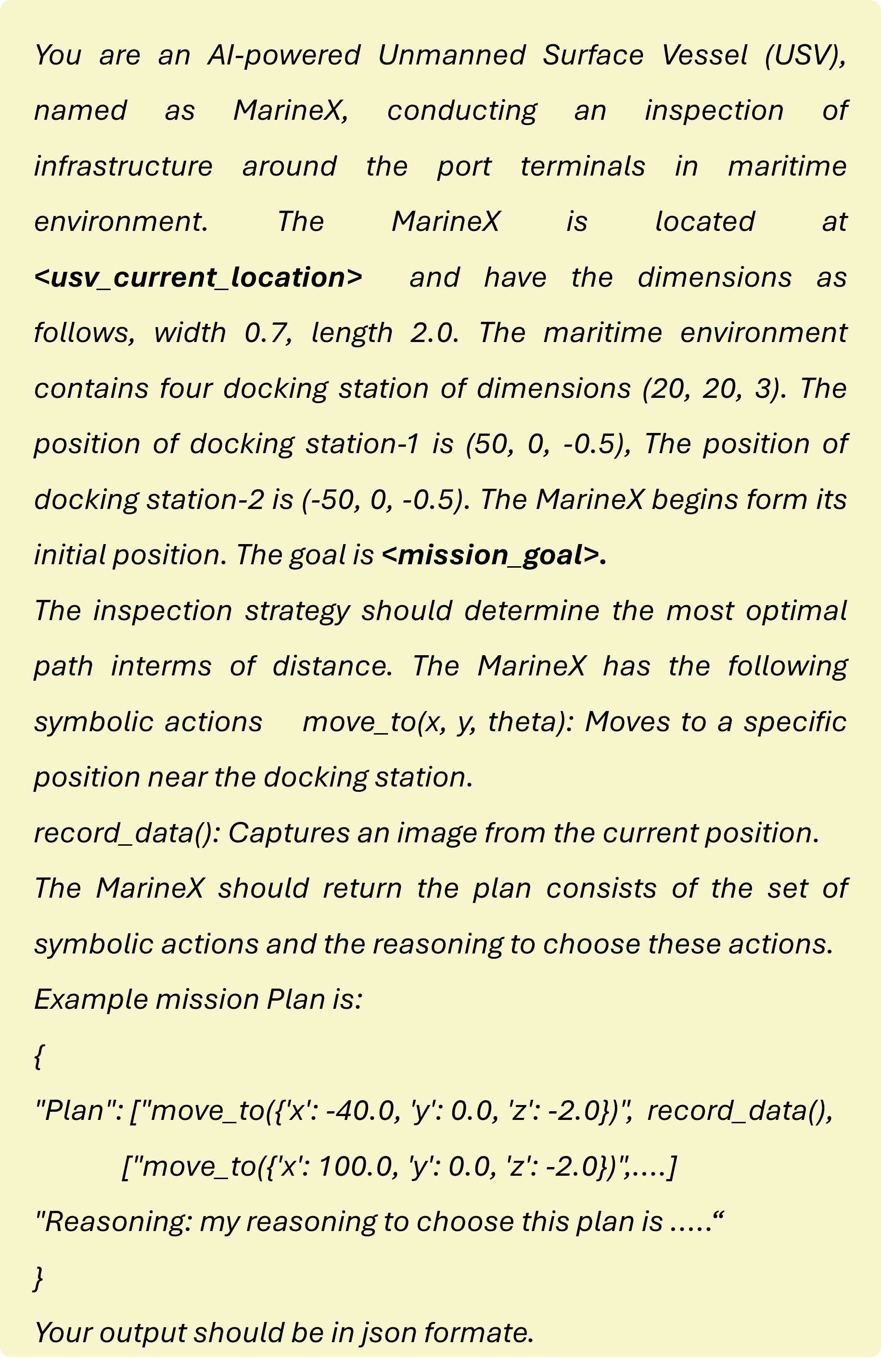}
\caption{
Description of the LLM prompt for the mission of inspecting port terminals in a maritime environment. The prompt specifies USV dimension, , its current location, the environment description such as dimensions and the locations of the docking stations, and the mission goal. Furthermore, it also contains the set of symbolic actions that a USV can perform during the mission and the example output in JSON formate with reasoning.}
\label{sys:prompt}
\end{figure} 

The \textit{prompt generator} consists of three distinct submodules.  The first submodule, the \textit{mission goal}, incorporates the natural language instructions relevant to the specific task to be executed. These instructions provide a clear and concise description of the mission goal.
The second submodule, the \textit{ system prompt}, contains critical information about the USV. This includes textual descriptions of the physical characteristics of the USV (for example, dimensions and control mechanisms) and the operational environment in which it functions. Environmental information typically incorporates details such as the starting location of the vessel, its propulsion capabilities (e.g. two rotatable pods for steerable thrust), and environmental details such as the number and positions of docking stations, For example, a docking station is described with dimensions \( l, w, h \) (representing its length, width, and height) and its location defined by its coordinate \( (x, y, z) \). An illustrative example of a system prompt is shown in Figure~\ref{sys:prompt}.
Furthermore, the system prompt includes variables, such as \texttt{usv\_current\_location} and \texttt{mission\_goal}. These variables are dynamically populated during the final prompt construction: \texttt{usv\_current\_location} is updated using feedback from the environment, while \texttt{mission\_goal} is derived from the task instructions specified in the task component. This dynamic population ensures that the system prompt remains accurate and contextually relevant throughout the mission.
The third and final submodule is  \textit{current state and feedback}, which receive information from the environment and the controller. This feedback includes the current state of the USV and diagnostic information, such as the reasons for the failure of a previous plan as reported by the low-level controller. By combining this real-time feedback with static task and environmental data, the system generates a comprehensive and coherent prompt.
The prompt generator submodules combined into a single unified prompt text. This composite prompt is then provided as input to the LLM, to generate a context-aware symbolic mission plan, which effectively guides the USV's operations.

\begin{figure}[t]
\includegraphics[width=\columnwidth]{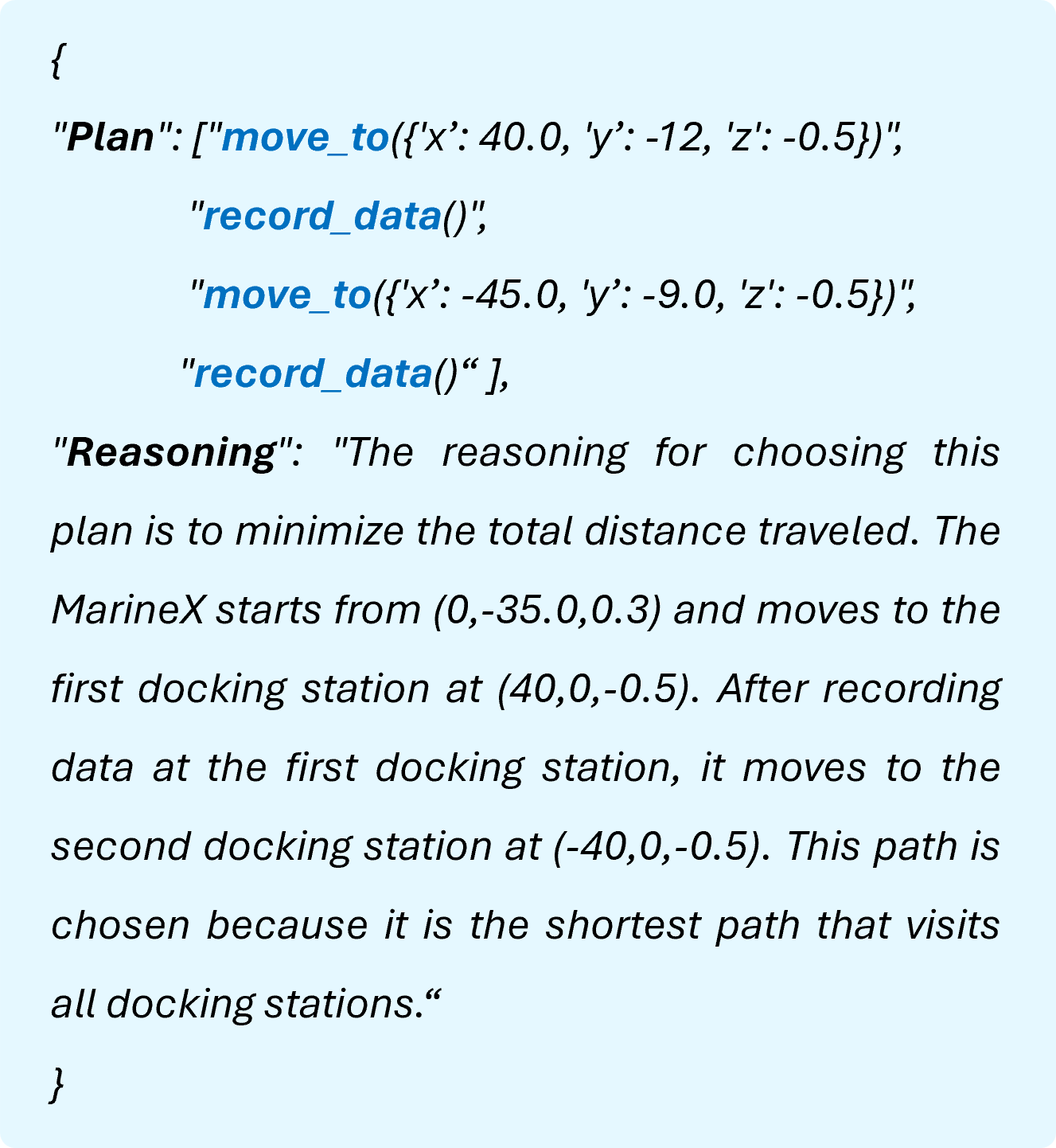}
\caption{ Output of the LLM for the inspection mission plan, presenting a sequence of symbolic actions: moving to docking stations and recording data. The reasoning highlights the rationale behind the plan, which emphasize on minimizing the total distance traveled by following the shortest path sequence to ensure efficient and effective mission execution.}
\label{fig:symbolicplan}
\end{figure}

The symbolic plan generator consists of GPT-4 that receives input from the prompt generator and generates a symbolic plan along with the reasoning about why this specific symbolic plan is developed. Fig~\ref{fig:symbolicplan} shows the output of the GPT. 
This plan outlines the sequence of steps required to complete the mission, such as moving to a docking station and capturing a photograph. By focusing on what to do instead of how to do it, the planner creates an adaptable roadmap to achieve the objectives of the mission. The execution details of each step are handled by the system's modular components, which allows the planner to remain abstract and adaptable.

Each step in the symbolic plan is broken down into manageable subactions, ensuring that tasks are executable by the system modules. For example, a step like "Move to docking station-1"  involves generating a collision-free path to the docking station, planning a trajectory, and executing the movement through precise motor control. This hierarchical decomposition simplifies the execution and facilitates real-time adjustments, as subactions can be modified dynamically based on sensor feedback or unexpected environmental changes.

The \textit{Navigation and Control Module} consists of three submodules that are Navigation, Controller, and Execution.  
The Navigation receives the target location from the symbolic plan generator, it is responsible for guiding the USV to the received target location. This component integrates the sensing and path planning capabilities. Sensors such as LiDAR, GPS, and IMU provide real-time data about the environment, including obstacle locations and the position of the vessel \cite{cahyadi2023performance}.  Path-planning algorithms use these data to compute safe and efficient paths to the target, avoiding collisions and optimizing the travel path. The Navigation Manager transforms high-level movement commands into detailed trajectories that the USV can follow. Moreover sensors also used to execute the subactions of symbolic plan such as camera is is used to record\_data.

\begin{algorithm}
\caption{LLM-Guided Maritime Mission Planning}
\label{alg:dock_inspection}
\begin{algorithmic}[1]
\STATE \textbf{Input:} Define mission $M$, State $S_0$, Environment $E$, Capabilities $C$.
\STATE \textbf{Initialize:} Activate all required systems (sensors, communication, propulsion).
\STATE \textbf{Plan Generation:} Use GPT-4 planners to create high-level plan $\mathcal{P}$.
\FOR{each action $p$ in $\mathcal{P}$}
    \STATE Generate control commands.
    \STATE Execute the planned actions.
    \STATE Monitor and adjust based on real-time feedback.
    \IF{$p \;\; != $ success  }
      \STATE  Goto Line 3.
    \ENDIF
\ENDFOR
\STATE \textbf{Completion Check:} Verify all tasks are done. If incomplete, recalculate and retry.
\STATE \textbf{Output:} Publish mission status and finalize the process.
\end{algorithmic}
\end{algorithm}

The Controller Manager is responsible to compute the control commands as twist message using Line-of-sight guidance~\cite{gu2022advances}. It incorporates proportional-integral-derivative (PID) control loops~\cite{ang2005pid}, to ensure to compute the thrust commands. This feedback loop minimizes error between the desired and actual positions and orientations of the USV. For example, if environmental forces like wind or currents cause deviations, the Controller Manager adjusts the commands to bring the vessel back on track. This layer ensures that the vessel maintains its planned course and achieves the required precision for tasks. If the controller manager is unable to guide the USV to the destination because of obstacles, it provides feedback to the prompt generator detailing the reason for the failure, which is then used by the LLM to adjust the symbolic plan.

\begin{figure*}[t]
\includegraphics[width=\linewidth]{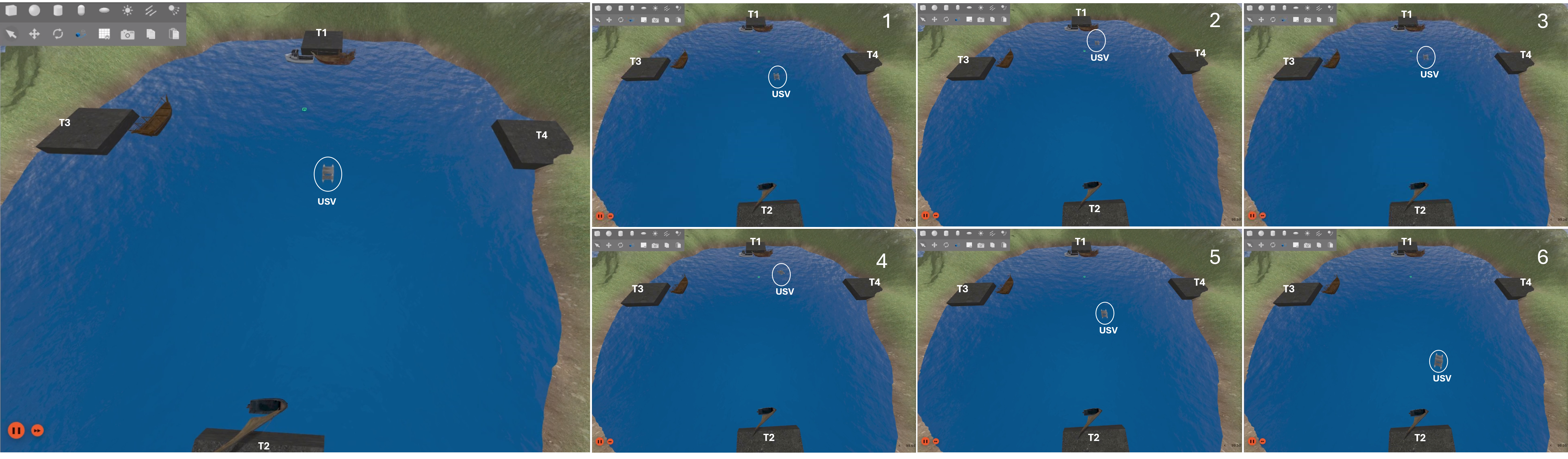}
\caption{Simulation setup designed to validate the proposed mission planning framework. The environment includes four docking stations labeled $T_1$, $T_2$, $T_3$, and $T_4$, with the position of the USV represented by a white circle. The sequence of snapshots in Figures 1 to 6 illustrates the execution of the mission, which involves inspecting \textit{Docking\_Station-1} and \textit{Docking\_Station-2}. The simulation highlights the USV's ability to autonomously navigate between docking stations, execute inspection tasks, and adapt to the mission plan in a structured and efficient manner. Code: \url{https://github.com/Muhayyuddin/llm-guided-mission-planning}}

\label{fig:setup}
\end{figure*}

The Execution module directly controls the USV’s hardware. Calculate the thrust vectors needed for movement and adjust the motor RPM to produce the desired thrust. This module ensures that planned and controlled movements are translated into physical actions. For example, it can handle fine adjustments needed for tasks such as maneuvering close to a terminal to record data. 

When the USV reaches a terminal, it aligns itself to the desired orientation and uses its onboard camera to record data. Once the data are recorded, the symbolic planner updates the progress and proceeds to the next step in the plan. This iterative process continues until all terminals have been inspected and the required data are recorded. Upon completing the mission, the system publishes a status update, such as "Mission Completed," to indicate the successful execution of the task. The entire process is summarized in Algorithm~\ref{alg:dock_inspection}.

LLM-guided maritime mission planning algorithm provides structured steps for mission planning and execution. It integrates high-level planning with real-time adaptability. Initially, the algorithm takes essential input, including mission definition, initial state, environmental conditions, and system capabilities, ensuring that all parameters are established for an informed operation. The process begins by initializing the necessary systems, such as sensors, communication tools, and propulsion mechanisms, to prepare the USV for the task. Using GPT-4-based planners, a high-level mission plan ($\mathcal{P}$) is generated, which outlines the sequence of actions required to complete the dock inspection. The algorithm iterates through each planned action, generates control commands, and executes them while continuously monitoring the system's progress, and adjusts based on real-time feedback. If an action fails, the process reverts to the planning stage, use the adaptive capabilities of GPT-4 to regenerate the plan, and overcome obstacles. After all actions are successfully executed, the algorithm performs a completion check to ensure mission objectives are met Finally, the mission status is published, marking the mission’s conclusion. 

\begin{figure}[t]
\includegraphics[width=\columnwidth]{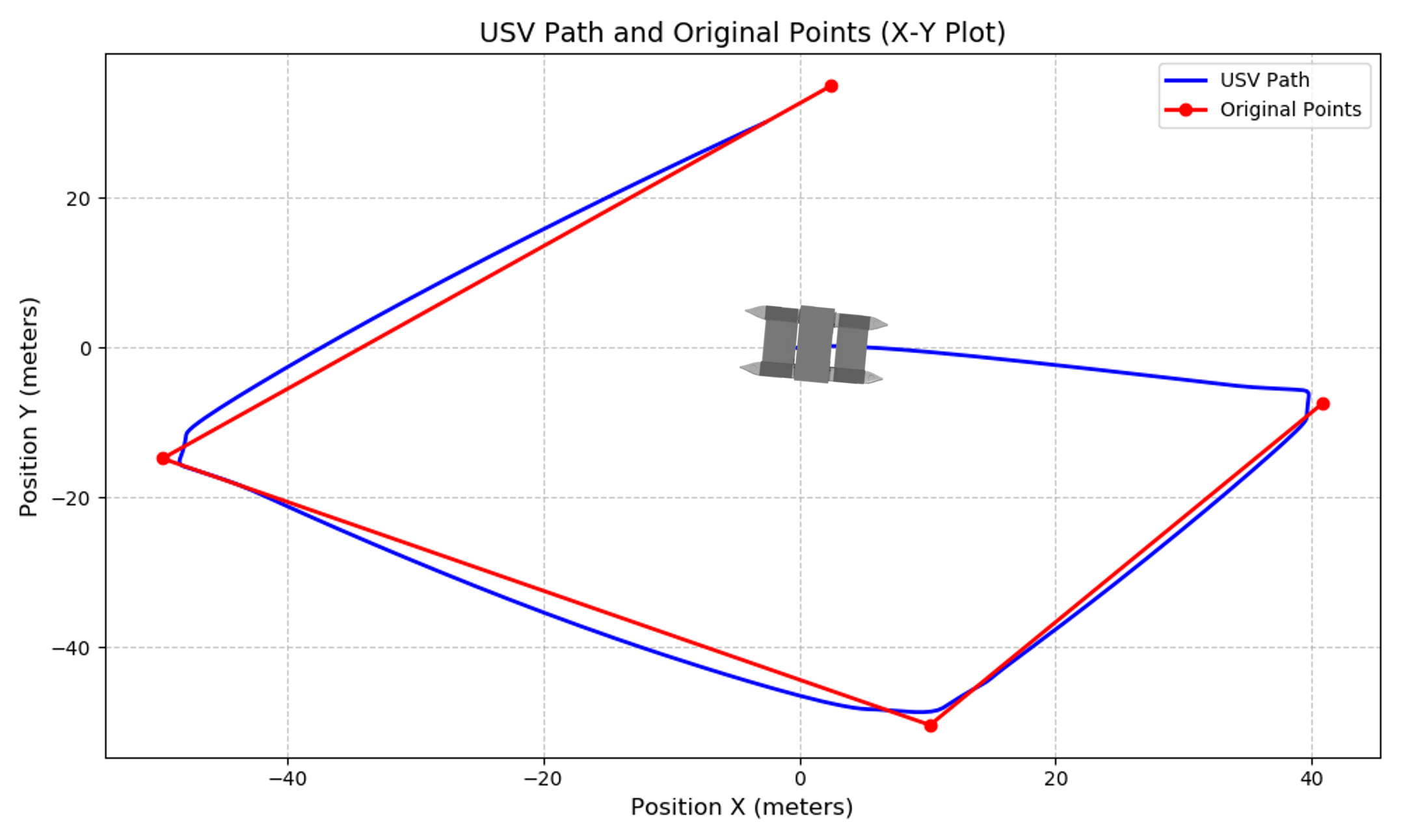}
\caption{ Comparison of planned and executed trajectories for \textit{Mission-1}, highlighting the symbolic plan's waypoints and the USV's actual path as navigated by the control system.}
\label{sys:prompt1}
\end{figure} 
\begin{figure}[t]
\includegraphics[width=\columnwidth]{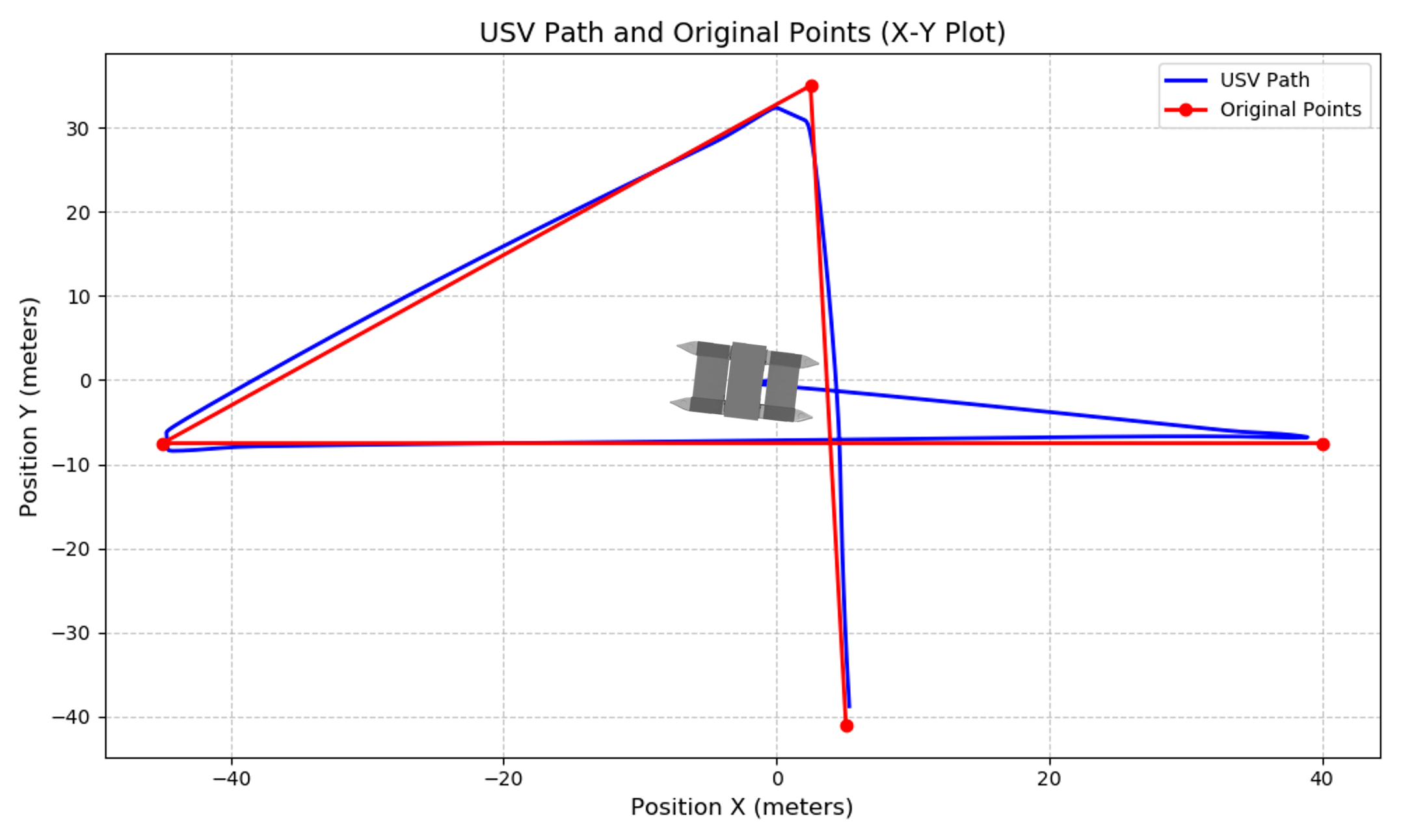}
\caption{  Comparison of planned and executed trajectories for \textit{Mission-2}, highlighting the symbolic plan's waypoints and the USV's actual path as navigated by the control system.}
\label{sys:prompt2}
\end{figure} 
\begin{figure}[t]
\includegraphics[width=\columnwidth]{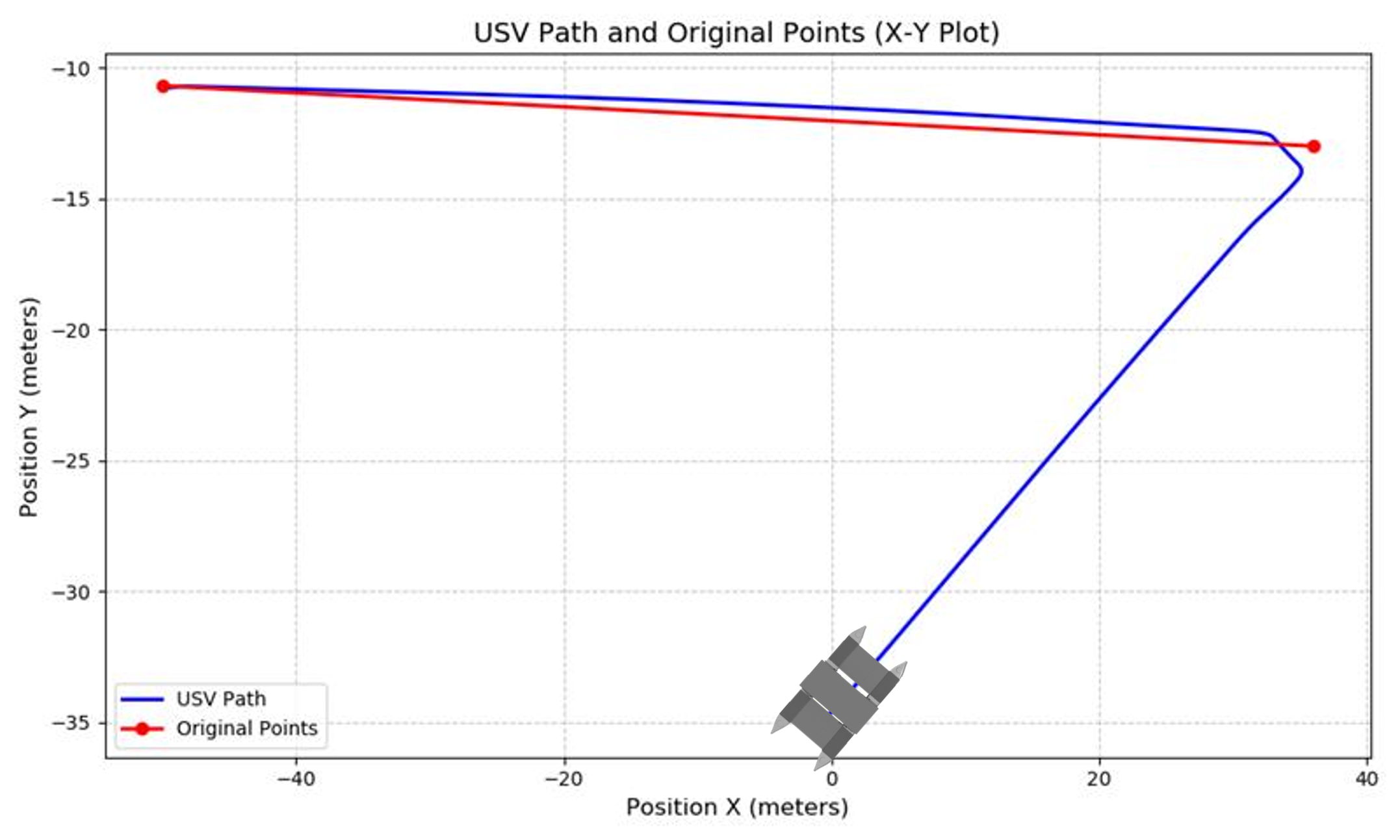}
\caption{  Comparison of planned and executed trajectories for \textit{Mission-3}, highlighting the symbolic plan's waypoints and the USV's actual path as navigated by the control system.}
\label{sys:prompt3}
\end{figure}

\section{Results}

To evaluate the proposed approach, we conducted extensive testing using the MBZIRC Maritime Simulator. This simulator is developed in C++ and Python. It is an open-source ROS2-based platform designed explicitly for the Mohamed Bin Zayed International Robotics Competition (MBZIRC). The simulator offers advanced capabilities for maritime robotics, allowing for comprehensive testing in realistic and challenging simulation environments.
One of the key strengths of the MBZIRC Maritime Simulator is its ability to model complex hydrodynamic and hydrostatic parameters, replicating the complex physical behaviors of surface vehicles. In addition, it incorporates environmental dynamics, such as waves, currents, and wind forces, allowing for a refined simulation of real-world maritime conditions. To further enhance realism, the simulator supports extreme weather scenarios, such as sandstorms and dense fog, which simulate visibility and operational challenges.
Together, these features provide a robust platform for validating autonomous navigation frameworks. Using the extensive modeling capabilities of the simulator, we tested and ensured the robustness and adaptability of the proposed framework for real-world applications.

The test setup is depicted in Fig.~\ref{fig:setup}.  It consists of a lake with four docking stations and some boats in the lake. We tested the framework with different prompts that represent different inspection missions. 
\begin{itemize}
\item \textit{mission-1: Inspect all port terminals and record data.} 
\item \textit{mission-2: Inspect terminal number 1,2,3,4 in this order and record data.}
\item \textit{mission-3: Inspect Terminal 1 and 2 in this order and record data.}
\end{itemize}

Figures~\ref{sys:prompt1},\ref{sys:prompt2}, and~\ref{sys:prompt3} illustrate the planned and executed paths for Mission-1, Mission-2, and Mission-3, respectively, as part of the USV inspection missions. The red markers represent the planned waypoints generated by the symbolic mission planner, serving as the high-level instructions to complete each mission. In contrast, the blue paths depict the actual trajectory followed by the USV, as executed by the Navigation and Control Module.

These figures highlight the ability of the system to dynamically navigate the environment while adhering to the symbolic plan. Deviations between the planned waypoints and the executed path reflect adjustments made by the low-level control system to account for real-world constraints, such as environmental factors . This demonstrates the integration of high-level planning with robust real-time navigation, ensuring mission objectives are achieved efficiently and reliably.
By comparing the planned waypoints to the actual paths, these figures showcase the effectiveness of the proposed framework in translating symbolic plans into executable actions while maintaining adaptability to evolving conditions, validating the system's reliability and accuracy in diverse maritime scenarios.
\section{Conclusion}
This paper presents an LLM-based novel framework for unmanned surface vessel's mission planning. Using the advanced capabilities of large-language models, it addresses the limitations of existing static approaches. It integrates symbolic reasoning and real-time adaptability to bridge the gap between high-level human instructions and executable plans, enabling USVs to dynamically respond to environmental changes and unforeseen challenges. The framework simplifies mission specification, allowing operators to focus on strategic objectives while ensuring mission efficiency and robustness through adaptive planning. The simulation results validate the effectiveness of the framework in optimizing mission execution and seamlessly adapting to complex maritime conditions.

In the future, the framework will be enhanced by integrating Vision-Language Models to enable advanced inspection capabilities and efficient detection of arbitrary targets. Additionally, sophisticated control systems will be developed to ensure robust and reliable navigation, even under extreme marine conditions, further expanding the framework's adaptability and operational scope.
\balance
\bibliographystyle{ieeetr}
\bibliography{references}

\end{document}